\documentclass{article}

\usepackage{arxiv}

\usepackage[utf8]{inputenc} % allow utf-8 input
\usepackage[T1]{fontenc}    % use 8-bit T1 fonts
\usepackage{hyperref}       % hyperlinks
\usepackage{url}            % simple URL typesetting
\usepackage{booktabs}       % professional-quality tables
\usepackage{amsfonts,amsmath}       % blackboard math symbols
\usepackage{nicefrac}       % compact symbols for 1/2, etc.
\usepackage{microtype}      % microtypography
\usepackage{lipsum}
\usepackage{graphicx}
\graphicspath{ {./images/} }

\title{Privacy-Preserving Data Linkage \\ Across Private and Public Datasets \\ for Collaborative Agriculture Research}

\author{
Osama Zafar \\
Department of Computer and Data Sciences\\
  Case Western Reserve University \\
  \texttt{oxz23@case.edu} \\
   \And
 Rosemarie Santa González \\
  H. Milton Stewart School of Industrial and Systems Engineering\\
  Georgia Institute of Technology\\
  \texttt{rgonzalez308@gatech.edu} \\
  \And
 Gabriel Wilkins \\
  College of Letters \& Science\\
  University of Wisconsin–Madison\\
  \texttt{gwilkins@wisc.edu} \\
  \And
 Alfonso Morales \\
  College of Letters \& Science\\
  University of Wisconsin–Madison\\
  \texttt{morales1@wisc.edu} \\
  \And
 Erman Ayday \\
  Department of Computer and Data Sciences\\
  Case Western Reserve University \\
  \texttt{exa208@case.edu} \\
}

\begin{document}

\maketitle

\begin{abstract}
Digital agriculture—the use of technology in agricultural practices—plays an integral role in agricultural research, with the potential to improve crop yield, disease resilience, and long-term soil health. However, it raises privacy concerns and risks, such as adverse pricing, price discrimination, increased insurance costs, and manipulation of essential agricultural resources. These concerns deter farm owners and operators from sharing their data because it could be used against them. This study addresses the need for privacy-preserving frameworks that enables secure data sharing for digital agriculture practices. We propose a privacy-preserving framework designed to mitigate privacy risks while enabling comprehensive data analysis. The proposed tool allows the agricultural community to benefit from research driven policy making that establishes meaningful connections across public and private digital agriculture datasets. Our privacy-preserving algorithm behind the framework can be used to: (i) identify similar farmers based on specified attributes using multiple private datasets, (ii) provide aggregate information such as time and location, (iii) determine price and availability trends of target products, and (iv) correlate these trends with relevant public policy databases, such as food insecurity data, without compromising privacy. We validate the framework using real-life datasets from Farmer’s Market. Also, we demonstrate the framework's efficacy and practicality by training various machine learning models on linked privacy-preserved private and public data. The numerical experiments showcase how the framework can support policymakers and researchers in implementing digital agriculture related to food insecurity and pricing. Our work addresses critical privacy concerns, thereby facilitating more effective and secure agricultural research. It represents a significant contribution to digital agriculture by enabling the secure integration and analysis of data, paving the way for transformative advancements in the use of technology for agricultural development.
\end{abstract}

% keywords can be removed
Code Repository: \href{https://github.com/ICICLE-ai/Collaborative_Agriculture_Research}{GitHub Collaborative Agriculture Research Repository}

\section{Introduction} \label{section:intro}

The importance of agriculture to human societies has driven researchers to improve cultivation and harvest methods by, for instance, creating fertilizers and pesticides, genetically modifying disease-resistant seeds, and assessing soil composition for crop fitness. This journey of innovation is now fueled by a constant stream of data harvested from satellites, unmanned aerial vehicles, self-driving vehicles, and sensors in the field. Although farmers occasionally share sales data as part of marketing research, the sharing of data related to their fields’ environmental and soil conditions can be more fraught with privacy concerns. The same value that such data provide to digital agriculture research gives extra weight to the potential misuse of such data, including adverse pricing, price discrimination, interference of potential diseases, hiked insurance costs, and negative interactions between farmers \cite{wiseman2019farmers, taylor2018climate}. Such risks can make farmers reluctant to share their data with outside researchers and institutions,  and thus a key focus on future digital agriculture research should include protecting farmers from these risks. Similar technique has been used in other fields, such as bioinformatics, for privacy-preserving identification of population stratification for genomics research \cite{Dervishi}.

This paper highlights the need for a privacy-perserving mechanism. Hence, we propose a framework designed to address the data safety and privacy concerns associated within digital agriculture. Our work aims to mitigate the potential misuse of agricultural data, which can lead to adverse outcomes and unfair practices in the market, while making farmers feel safe sharing their data with external researchers and institutions.

Our contributions are twofold. First, we develop and propose techniques that ensure the protection of farmers' data while maintaining its utility for meaningful analysis. By anonymizing data and implementing robust privacy protection, our framework safeguards against adverse pricing practices and protects farmers' personal information from exploitation. Second, we demonstrate how our framework can be utilized to analyze relationships between various datasets without compromising privacy. For instance, a researcher who is interested in exploring the relationship between the price of a particular product that is grown by farmers with some specific attributes can utilize our proposed techniques to safely identify farmers (from anonymized data) who match a profile based on those particular attributes. Subsequently, the researcher can query pricing information from these selected farmers over a defined period, which can be used to establish an association between product pricing and food insecurity, using the timeline as a common denominator for analysis.

The remainder of this study is structured as follows. To lay the groundwork for our proposed solutions, we start with a literature review of foundational concepts and related works in \textbf{Section \ref{section:background}: Background and Related Work }. Next, \textbf{Section \ref{section:solution}: Proposed Solution} delves into the design and technical details of our framework. Then, we examine the privacy and utility evaluation metrics and performance of our framework based on the said metrics in the \textbf{Section \ref{section:eval}: Privacy and Evaluation}. \textbf{Section \ref{section:applications}: Applications in Machine Learning} explores the application of our framework to real-world scenarios and demonstrates the practical utility of our approach. Finally, the \textbf{Section \ref{section:conclusion}: Conclusion} summarizes our contributions and discusses their potential impact.

\vspace{1em}
\vspace{1em}
\section{Background and Related work} \label{section:background}
This sections explores key areas relevant to our study, including data privacy challenges in smart agriculture, existing frameworks and mechanisms for secure data sharing, and data privacy techniques, each providing critical insights into the current state of research and the gaps our work aims to address.

\vspace{1em}
\subsection{Agricultural Privacy Concerns}
As the agricultural industry evolves, farmers are increasingly adopting technologies to manage soil, water, and harvesting more efficiently. Since 2010, these technologies, collectively known as Climate Smart Agriculture (CSA) or precision agriculture, have driven advancements in large-scale farming \cite{FAO}. These practices are vital for commodity farmers, enhancing resource use and sustainability by precisely allocating inputs according to crop conditions and, yet, can exacerbate inequalities\cite{taylor2018climate}.

Small and medium-sized farmers are hesitant to adopt CSA technologies for two main reasons. First, these technologies are better suited to large mono-crop operations, making them less applicable to diversified farms \cite{WAKWEYA2023100698}. Second, large-scale farmers often share data with industry partners to receive tailored advice, but diversified farmers fear their data may be misused, leading to privacy concerns \cite{taylor2018climate, WAKWEYA2023100698, WESTERMANN2018283}. Since 2015, academic research has increasingly addressed farmers' data privacy concerns, suggesting solutions such as legislative protection and software systems \cite{linsner2021role, Kaur2022}. However, as of August 2024, there is no federal body specifically focused on data protection for agriculture, unlike the healthcare and finance sectors, which have regulatory agencies like HIPAA and GBLA \cite{ferris2017data}. The U.S. Cybersecurity and Privacy Operations Center (CPOC) exists \cite{USDA} but lacks the same authority to enforce policies across the agricultural network \cite{ferris2017data}.

Agricultural data storage involves various entities, including Agricultural Technology Providers, farmers, local agronomists, and digital silos—software systems like Edge Computing and Blockchain that offer robust data protection \cite{amiri2022big}. Despite these efforts, small-scale farmers face challenges in adopting these complex technologies, bearing significant risks without seeing proportional benefits \cite{wiseman2019farmers}. Their reluctance to share data under lengthy, technical agreements further exacerbates this issue.

\vspace{1em}
%\subsection{Data Sharing Mechanisms for Agriculture}
\subsection{Data Sharing Frameworks for Digital Agriculture}
The literature on data sharing frameworks in digital agriculture has seen developments aimed at addressing privacy and security concerns. Spanaki and coauthors \cite{SPANAKI2021102350}, propose a set of principles and a released access control mechanism within Data Sharing Agreements (DSAs) between the agricultural sector and AI application users. This approach is designed to address privacy concerns by regulating access to shared data \cite{SPANAKI2021102350}. On the other hand, Song and coauthors \cite{9170612}, proposes a flexible, cloud-based privacy-preserving data aggregation scheme. This scheme allows users to access aggregated plain-text data from farmers without risking information leaks, offering a practical solution to maintaining data privacy in cloud environments \cite{9170612}.

However, most studies have concentrated on attacks or data breaches from smart devices. \cite{Gupta2020} propose a multi-layered architecture combined with cyber-physical environment measures to deter external cyberattacks on smart farming systems. This layered approach enhances the resilience of agricultural technologies against potential threats. Similarly, \cite{Kumar2022} present a privacy encoding framework tailored for sensors used in digital agriculture. Their work also includes an intrusion detection system to safeguard against unauthorized data access, ensuring that data integrity is maintained within agricultural networks \cite{Kumar2022}. \cite{KUMAR2021107819} introduce a secure privacy-preserving framework for unmanned aerial vehicles (UAVs). This framework focuses on data authentication processes that mitigate risks associated with data poisoning attacks, which could otherwise compromise the reliability of agricultural data \cite{KUMAR2021107819}.

There is a necessity for tailored security frameworks that specifically address the needs of these medium and small farming operations \cite{ongadi2024comprehensive}. To the best of our knowledge, there is dearth of frameworks that facilitate effective data collection from farmers. Our study addresses this gap by proposing and testing a framework that integrates data collected from farmers with publicly available datasets, making it suitable for use in machine learning (ML) and AI applications. Our framework not only ensures the privacy of individual farmers but also enhances the utility of the aggregated data for broader AI-driven agricultural research.

\vspace{1em}
\subsection{Data Privacy Techniques}
In the broader context of data-driven approaches, protecting sensitive information is crucial across various domains, including agriculture. While some data privacy techniques have been widely applied in fields such as finance, and healthcare, they also hold potential for enhancing data security in digital agricultural applications. This section examines several key data privacy techniques, exploring both their theoretical foundations and their adaptability for use in agriculture. We will discuss Principal Component Analysis (PCA), which, though not explicitly designed for privacy, offers a form of data abstraction; Local Differential Privacy (LDP), a method that ensures robust privacy through local data perturbation; and K-means clustering, a widely used algorithm that can be adapted for privacy-preserving purposes. Understanding the mathematical principles behind these techniques is essential to appreciate their role in safeguarding sensitive data, particularly as the proposed framework is presented.

\textit{\textbf{Principal Component Analysis (PCA)}} \cite{Pearson} is a widely utilized technique for data analysis and the identification of patterns and relationships within data. It has a wide variety of applications, such as image compression \cite{Gaidhane}; facial recognition \cite{Gottumukkal}; medical data correlation \cite{Qureshi}; and Quantitative Finance \cite{Yu}. PCA itself does not inherently provide any guarantees of privacy, as its primary function is to reduce the dimensionality of data by transforming it into a set of orthogonal components. However, this transformation abstracts the data, making it harder to reverse-engineer individual records and thereby providing a form of data obfuscation.

\textit{\textbf{Local Differential Privacy (LDP)}} \cite{Duchi, Kairouz} is a privacy-preserving technique method that builds upon the principles of classic Differential Privacy (DP) \cite{Dwork}. LDP is designed to protect individual data and extends the concept of DP to a decentralized context. In LDP, each user perturbs their own data prior to sending it to a data collector. This local perturbation ensures that the privacy of individual data is protected, even in cases where the data collector is not considered reliable. Unlike DP, which relies on a central trusted party that has access to the raw data, LDP offers privacy assurances directly at the data source. An algorithm \( A \) is considered to satisfy \( \epsilon \)-LDP if, for any pair of individual private data points \( a_1 \) and \( a_2 \), and for any possible output \( b \),:

\begin{equation}
\label{equation:1}
\Pr[A(a_1) = b] \leq e^\epsilon \cdot \Pr[A(a_2) = b],
\end{equation}

where \( \epsilon \) represents the privacy parameter. \( \epsilon \)-LDP is achieved by introducing an appropriate amount of noise to individual data points. The primary challenge is determining the noise's magnitude to ensure compliance with LDP while preserving data utility. Various mechanisms have been developed in the DP field to address this problem, with the Laplacian mechanism \cite{Dwork2006} being one of the most prominent.

For any numerical function \( f(x): \mathbb{R} \rightarrow \mathbb{R} \) that satisfies \( \epsilon \) -LDP if Laplacian noise is added as follows:

\begin{equation}
F(x) = f(x) + Lap\left(\frac{s}{\epsilon}\right)
\end{equation}

where \( s \) represents the sensitivity of the function \( f \) and \( \text{Lap}(k) \), where \( k = \frac{s}{\epsilon} \), denotes sampling from a Laplace distribution with scale \( k \). Sensitivity measures the maximum change that a single data point can cause to the output of \( f \) in the worst-case scenario. In our work, we calculate sensitivity to add Laplacian noise \cite{Dwork2013}. By definition, \( s = \max_{x, x'} \|f(x) - f(x')\|_1 \).

\textit{\textbf{Kmeans}} clustering algorithm \cite{Hartigan} is a widely used unsupervised machine learning technique designed to partition a dataset into \( K \) distinct, non-overlapping clusters. The primary objective of K-means is to minimize the intra-cluster variance, thereby ensuring that data points within the same cluster exhibit high similarity, while data points in different clusters display significant dissimilarity. Mathematically objective function to be minimized is as follows:
\[
J = \sum_{j=1}^{K} \sum_{x_i \in C_j} \| x_i - \mu_j \|^2
\]
where \( \| x_i - \mu_j \|^2 \) is the squared Euclidean distance between data point \( x_i \) and centroid \( \mu_j \) of cluster \( C_i\). To determine the optimal number of clusters in the Kmeans clustering data, we employ the Elbow method \cite{Syakur}. This technique involves plotting the explained variation, measured as the within-cluster sum of squares (WCSS) (the sum of the square distance between points in a cluster and the cluster center), against a range of possible cluster numbers. The WCSS quantifies the variance within each cluster, with lower values indicating tighter and more compact clusters. By examining this plot, we look for a point where the rate of decrease in WCSS sharply shifts, forming an elbow. We pick the number of clusters corresponding to this elbow as optimal.

\vspace{1em}
\vspace{1em}
\section{Proposed Solution} \label{section:solution}

The proposed framework works as a centralized sandbox (\textit{i.e.,} an isolated environment) where researchers can safely interact with both private digital agriculture data from farmers' markets, each containing data from multiple farmers, and publicly available digital agriculture datasets. 

\begin{table}%[h!]
\centering
\caption{Notations and descriptions}
\label{table:symbols}
\vspace{1em}
%\resizebox{\columnwidth}{!}{
\begin{tabular}{ll}
\hline
\textbf{Notation} & \textbf{Description} \\
\hline
$R$ & Researcher \\
$F_i$ & Farmer's market $i$ \\
$D_s$ & Dataset used by the researcher to train the PCA model \\
$M_s$ & Trained PCA model \\
$O_i$ & Original PCA matrix of framer $F_i$ \\
$C_i$ & Noisy PCA matrix of framer $F_i$ \\
\hline
\end{tabular}
%}
\end{table}

\begin{figure}[t]
    \centering
    \includegraphics[width=.7\linewidth]{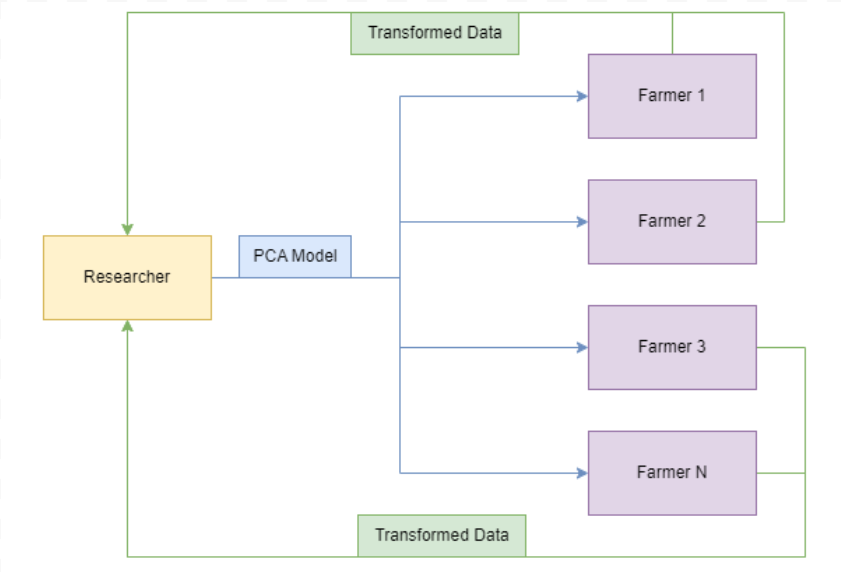}  
    \caption{Illustrates a high-level architecture of the framework: Researcher sends a trained PCA model to farmers, who then transform their data and share it back with the researcher.}
    \label{fig:figure1}
\end{figure}

\begin{figure*}[!t]
  \includegraphics[width=\textwidth,height=10cm]{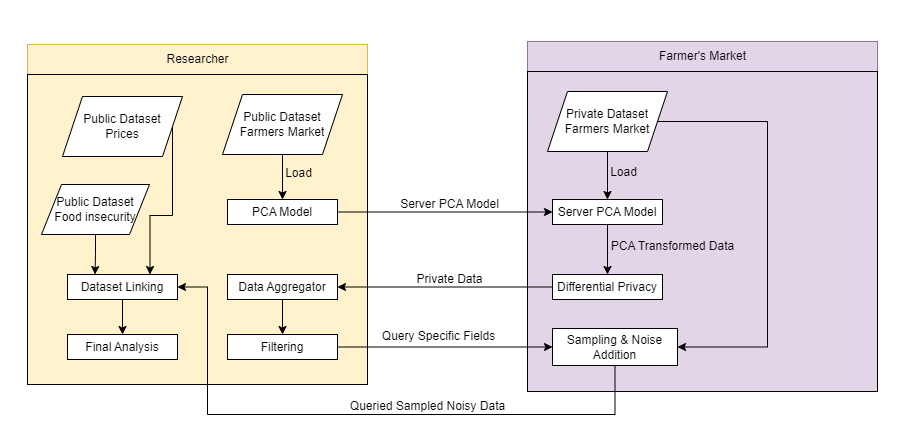}  
  \caption{Detailed architecture of the framework.}
  \label{fig:figure2}
\end{figure*}

We propose a privacy-preserving framework based on dimensionality reduction to identify farmers (\textit{i.e.}, records in the private datasets) with a predefined set of attributes from private datasets. 
See Table~\ref{table:symbols} for the list of notations and their definitions. 
Let $R$ denote the researcher and $F_i$ denote the farmer's market $i$. As illustrated in Figures \ref{fig:figure1} and \ref{fig:figure2}, the researcher $R$ trains a global principal component Analysis (PCA) model on a publicly available dataset $D_s$ that has desired features. We use StandardScaler from scikit-learn \cite{pedregosa} to preprocess $D_s$. This trained PCA model $M_s$ is then shared with all the farmer's markets along with the sensitivity parameter, which will be used later in the noise addition step. Each $F_i$ uses model $M_s$ to transform its private data into PCA matrix $O_i$. In order to guarantee the privacy of the data, especially against membership inference attacks, the framework adds Laplacian noise locally to achieve $\epsilon$-LDP. Noise is added to each component of a PCA matrix using \ref{equation:1} where sensitivity $s$ is replaced with $s_i$, which denotes the sensitivity of component $i$ in the PCA matrix. Finally, a differentially private version of their PCA output $C_i$  is shared with researcher $R$. The researcher $R$ combines the received PCA outcomes from multiple datasets and identifies the desired farmers accordingly. Note that during this process, the researcher $R$ does not have direct access to the private datasets, and this identification of the desired set of farmers is done only by using differentially private PCA outputs from the farmers. 

Once the researcher identifies the farmers with the desired set of attributes across different private datasets, then the researcher sends queries to the corresponding farmer markets through the sandbox environment in order to receive aggregate information only for the selected set of farmers. Such aggregate information is then used by the researcher to find relationships between these farmers’ data and those from public digital agriculture datasets. 

Figure \ref{fig:figure2} illustrates the complete workflow of the framework, with an example of datasets linking between a private farmers market dataset and a public food insecurity dataset. Initially, the researcher requests a privately formatted farmer dataset, and this is then followed by the aforementioned privacy-preserving processes to identify farmers with the desired characteristics. Subsequently, relevant features are queried to facilitate dataset linking and analysis.

Furthermore, the framework allows researchers to explore private datasets in a privacy-preserving manner and link private and public datasets to expand the scope of their data collection, investigate correlations, and analyze instances requiring sensitive and private agricultural data. For example, a researcher aims to study the effect of produce price on its sale. The researcher collects data from farmer markets – in a privacy-preserving way using the proposed framework – and applies clustering algorithms to identify closely related groups. Specifically, the researcher employs the KMeans clustering algorithm \cite{Hartigan} on a PCA-transformed version of a public dataset combined with the previously gathered private datasets. The KMeans labels are then used to group farmers and recommend connections within these groups. 

\vspace{1em}
\subsection{Case Study:}
Privacy-preserving data sharing unlocks a world of applications in various domains, from healthcare research to financial analysis. For the purpose of feasibility testing and evaluation, we consider the following use case from the domain of digital agriculture research: The research seeks to analyze the relationship between the sales of a specific product and its market price over time, while also investigating how these factors correlate with food insecurity.

\begin{figure}[!t]
    \centering
    \includegraphics[width=\linewidth]{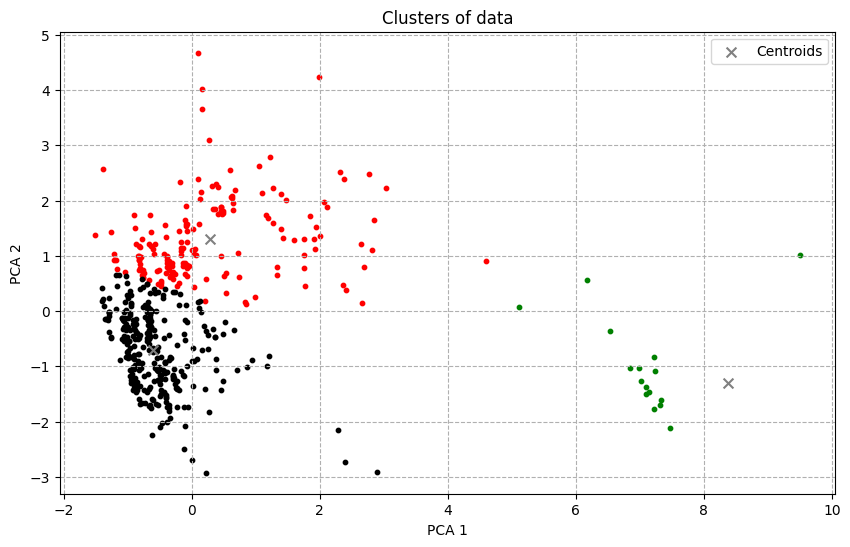}  
    \caption{PCA Transformed Farmer's Market Dataset Clustering.}
    \label{fig:figure3}
\end{figure}

% \begin{figure*}
%   \includegraphics[width=\textwidth,height=5cm]{./Images/Figure0.png}
%   \caption{Farmers Market Dataset.}
%   \label{fig:figure0}
% \end{figure*}

% ID	Miles from Market - Primary Location	Fruits & Vegetables - Vendor Type	Meat & Seafood - Vendor Type	Dairy - Vendor Type	Eggs - Vendor Type	Plants & Flowers - Vendor Type	Nuts & Legumes - Vendor Type	Value-added - Vendor Type	Prepared Food - Vendor Type	Crafts/Art/Services - Vendor Type	Sales	Visitor Count
% 6815	42.22	0	0	1	0	0	0	1	1	0	414.0	70
% 5991	19.62	1	1	0	0	0	0	0	0	0	157.0	404
% 5663	23.54	0	0	0	0	0	0	0	0	1	670.0	6
% 5950	70.52	0	0	0	0	0	0	1	0	0	70.0	53
% 5686	0.68	0	0	0	0	0	0	0	0	0	285.0	166

\begin{table*}[!t]
\centering
\caption{Farmers Market Dataset}\label{table:table2}
\vspace{1em}
%\resizebox{\textwidth}{!}{
\begin{tabular}{cc c c c c c c c c c rr}
\toprule
    & \textbf{Miles from Market} 
    & \multicolumn{9}{c}{\textbf{Vendor Type}}
    & \textbf{} \
    & \textbf{} \\
\cmidrule(lr){3-11}
    \textbf{ID} 
        & \textbf{Primary Location} 
        & \textbf{F\&V} 
        & \textbf{M\&S} 
        & \textbf{D} 
        & \textbf{Eggs} 
        & \textbf{P\&F} 
        & \textbf{N\&L} 
        & \textbf{VA} 
        & \textbf{PF} 
        & \textbf{CAS} 
        & \textbf{Sales} 
        & \textbf{\#Visitors} \\ 
\midrule
6815 & 42.22 & 0 & 0 & 1 & 0 & 0 & 0 & 1 & 1 & 0 & 414.0 & 70 \\
5991 & 19.62 & 1 & 1 & 0 & 0 & 0 & 0 & 0 & 0 & 0 & 157.0 & 404 \\ 
5663 & 23.54 & 0 & 0 & 0 & 0 & 0 & 0 & 0 & 0 & 1 & 670.0 & 6 \\ 
5950 & 70.52 & 0 & 0 & 0 & 0 & 0 & 0 & 1 & 0 & 0 & 70.0 & 53 \\ 
5686 & \phantom{0}0.68 & 0 & 0 & 0 & 0 & 0 & 0 & 0 & 0 & 0 & 285.0 & 166 \\
\bottomrule
\end{tabular} \\
{\footnotesize F\&V: Fruits and Vegetables; M\&S Meat and Seafood; D:Dairy; P\&F: Plants and Flowers; N\&L: Nuts and Legumes;\\ VA: Value-added; PF: Prepared Food; CAS: Crafts/Art/Services}
%}
\end{table*}

\begin{figure}[t]
    \centering
    \includegraphics[width=.5 \linewidth]{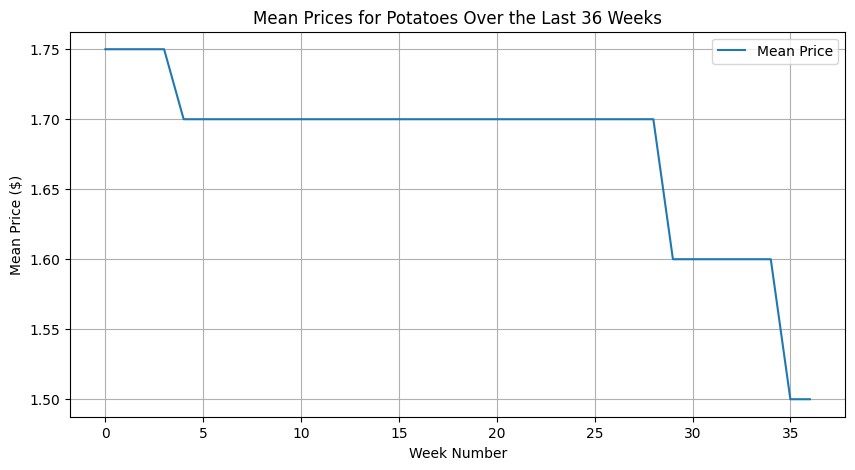}  
    \caption{Price of Potatoes in 2018.}
    \label{fig:figure4}
\end{figure}

%The dataset under consideration is the Wisconsin Farmer’s Market dataset (Figure \ref{fig:figure0}).
The dataset under consideration is a Farmer’s Market dataset (Table \ref{table:table2}).
%The Wisconsin Farmer's Market dataset was gathered internally by the University of Wisconsin. 
This data set was gathered internally and has not been made public. It includes detailed information about various aspects of farmer's market activities and vendors, 
%in Wisconsin, 
and it was collected specifically for research purposes. Due to the private nature of the information contained in this dataset, it remains private. The dataset is divided into multiple pieces representing various markets collecting data from different farmers and sharing it with researchers. We consider a researcher receives the dataset in a preserving manner using the above-mentioned technique and applies clustering. Figure \ref{fig:figure3} illustrates the plot of KMeans clustering on the PCA-transformed dataset. This approach allows researchers to identify farmers who closely match a desired profile for further research. The analysis involves a heterogeneous combination of datasets with varying levels of privacy implications, which could include public-private or private-private dataset combinations. For example, the researcher can combine a publicly available time series produce pricing dataset, like Fruits and Vegetables Prices Dataset \cite{Pricing} with the private 
%Wisconsin 
Farmer’s Market dataset to analyze produce prices comprehensively. Figure \ref{fig:figure4}: showing the average pricing of potatoes in 2018. Let's consider one of the groups in Figure \ref{fig:figure3} matches the generic profile of the farmers producing potato crop. The researcher can request sales data for these particular farmers for a particular time frame. The sales data can then be linked to the potato prices using time as a common reference, allowing to investigate the correlation between potato prices and sales. Figure \ref{fig:figure6} illustrates a plot of potato prices and sales of 2 farms over 36 weeks, reflecting their correlation. 

\begin{figure}[t]
    \centering
    \includegraphics[width=\linewidth]{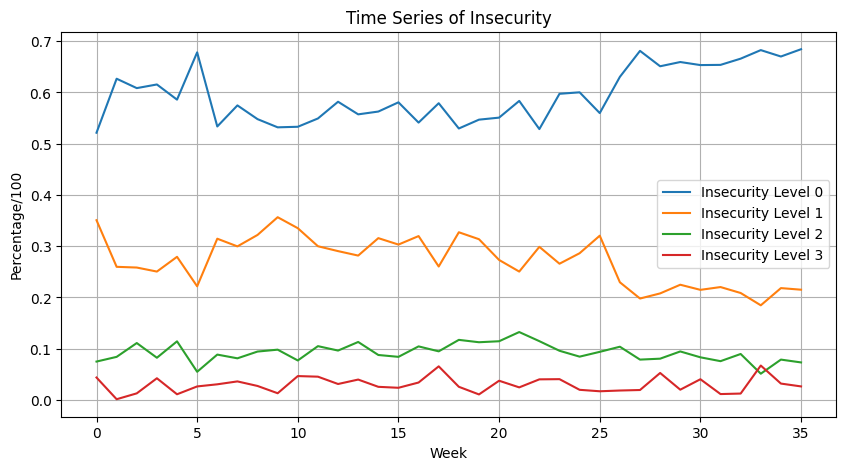}  
    \caption{Insecurity level measured in population (percentage) over time.}
    \label{fig:figure5}
\end{figure}

Similarly, the researcher can consider the food insecurity dataset \cite{FoodInsecurity} (Figure \ref{fig:figure5}: showing various levels of food insecurity over a timeline) and correlate it with sales data over the same timeline to study the impact of food availability and accessibility indicated by market sales and prices on food insecurity. 

\begin{figure}[t]
    \centering
    \includegraphics[width=\linewidth]{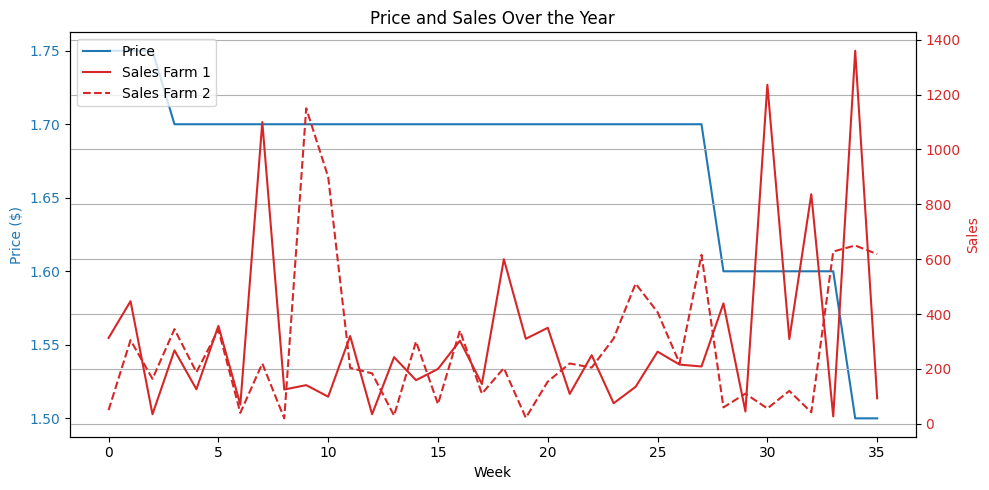}  
    \caption{Price and sales of potato recorded over time.}
    \label{fig:figure6}
\end{figure}

\vspace{1em}
\vspace{1em}

\section{Privacy and Utility Evaluation} \label{section:eval}

We consider honest farmer's markets that have legitimate datasets and follow the protocol. On the other hand, we consider an honest but curious researcher. In other words, the researcher correctly conducts the computations and procedures
but may try to infer sensitive information (about the individuals in farmer's market datasets) from the data sent to the researcher. The most well-known attacks on such datasets include the membership inference attacks. In the aforementioned attack, the adversary(researcher) has access to a victim's data, i.e., farmer, and tries to determine from the provided data whether a victim is a
part of one of the farmer's market's datasets. In our work, the framework aims to protect the privacy of the participants by ensuring that the researcher only learns the information explicitly provided and no additional details.

We consider the Wisconsin Farmer's Market dataset (See Table \ref{table:table2}) from our case study section. To simulate two local farmer's market datasets $F_i$s and the dataset used by the researcher to train the PCA model $D_s$, we divide our dataset into three parts. Each $F_i$ converts their data to $C_i$ using shared PCA model $M_s$ and shares it with researcher $R$. 

\begin{figure}[t]
    \centering
    \includegraphics[width=.5\linewidth]{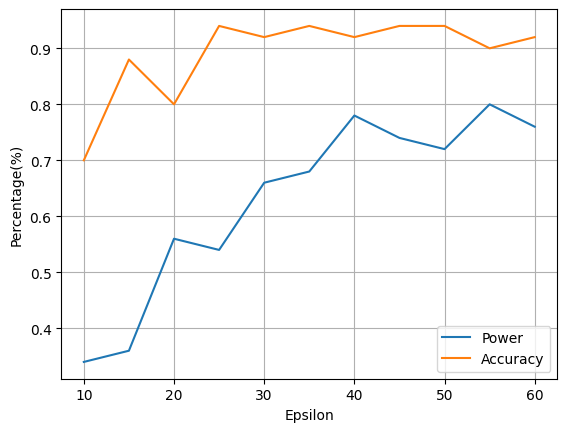}  
    \caption{Percentage for Noise Power and Accuracy (Logistic Regression) vs Epsilon value.}
    \label{fig:figure7}
\end{figure}

\begin{figure}[t]
    \centering
    \includegraphics[width=.5\linewidth]{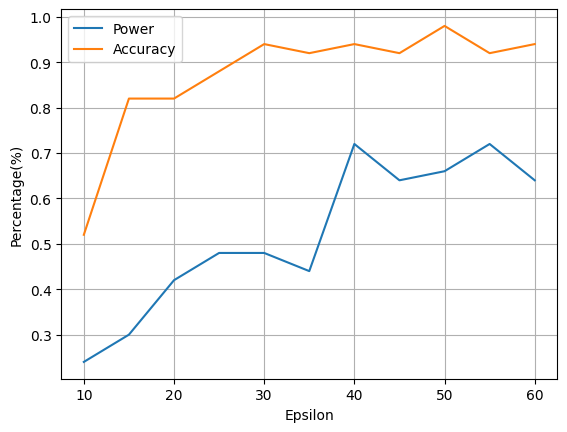}  
    \caption{Percentage for Noise Power and Accuracy (Naive Bayes) vs Epsilon value.}
    \label{fig:figure8}
\end{figure}

\begin{figure}[t]
    \centering
    \includegraphics[width=.5\linewidth]{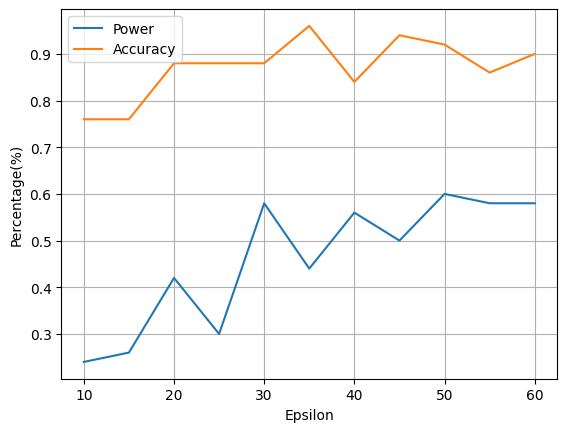}  
    \caption{Percentage for Noise Power and Accuracy (SVM) vs Epsilon value.}
    \label{fig:figure9}
\end{figure}

To evaluate the privacy performance of our algorithm, we compute the power metric, which indicates the fraction of correctly identified noisy samples. To achieve this, we calculate the minimum pairwise distances between the samples of the shared dataset and the control group as the control distances. Similarly, we compute the minimum pairwise distance of the shared dataset and case group as case distances. Here, the control group is sampled from the server's global dataset, and the case group is sampled from the shared dataset. Then, we determine the threshold, defined as the distance at the 95th percentile (when FPR is 5\%) of the sorted control distances. This threshold is subsequently used to evaluate the fraction of case distances that fall below it, thereby determining the power. The metric is evaluated with specific thresholds for False Positive Rate (FPR) and epsilon ($\epsilon$). Epsilon is a metric of privacy loss at a differential alteration in the dataset and is allocated across columns according to their sensitivity. For our analysis, we set the FPR at 0.05 (5\%) and incrementally vary the epsilon value, starting from 10, to observe its impact on the power metric. This approach helps us select the correct value of the epsilon with acceptable robustness represented by the power metric. For the utility metric, we compare the noise-free cluster labels of the data with the labels predicted by a binary classifier for noisy samples, assessing the accuracy of correctly predicting the labels of noisy samples. We have performed the aforementioned utility test using three different algorithms (Logistic Regression, Naive Bayes, and Support Vector Machine) to demonstrate the results across classifiers. Figures [\ref{fig:figure7}, \ref{fig:figure8}, \ref{fig:figure9}] illustrate the power and accuracy for a range of epsilon values for the Logistic Regression, Naive Bayes and SVM (Support Vector Machine) algorithms, respectively. The results reflect an acceptable robustness of privacy with good utility accuracy. For each model optimal epsilon value is selected for which we have low power and high accuracy and, therefore, optimal epsilon values are 25, 35, and 35 for Logistic Regression, Naive Bayes, and Support Vector Machine models respectively.

\vspace{1em}
\vspace{1em}

\section{Applications in Machine learning} \label{section:applications}
The proposed framework enables researchers to link private and public datasets, thereby expanding the scope of their research. It could also aid in training a diverse array of machine-learning models based on these newly identified correlations. We explore a few examples based on the case study dataset that researchers can use to train machine learning models: Researchers can train a model to predict sales based on factors such as prices, visitor count, and distance from the market. Additionally, they can link food insecurity datasets \cite{FoodInsecurity} with sales and pricing information to develop a model for predicting food insecurity or to study the impact of pricing on insecurity levels. Alternatively, they can train a machine learning model to predict food accessibility based on factors such as distance from the market, visitor logs, and insecurity levels.

In general, our proposed framework opens up a vast array of potential machine-learning applications. By combining the diverse data sources, researchers can uncover insights, patterns, and correlations that would be impossible to identify without access to the private datasets. Here are a few potential machine learning applications based on our case study dataset: linking soil condition data of farms with weather data to propose management practices and predict crop yields in the region; combining weather data with soil quality, soil moisture, and crop water requirements to train models for optimizing irrigation systems and predicting water needs in the region; and assessing the impact of climate change on crop yield and soil quality.
\vspace{1em}
\vspace{1em}

\section{Conclusion} \label{section:conclusion}

Privacy is a major concern when it comes to sharing sensitive data, especially if it could potentially be used against the data owner. This is very true for farmers, and as such, they are often hesitant to share their farm data with researchers. However, such farm data is helping to fuel innovation and research in digital agriculture, and thus a privacy-preserving data-sharing solution is vital to secure farmer data and allow researchers to safely utilize it. Our framework achieves this in an effective and efficient way using techniques such as PCA and differential privacy frameworks to ensure the individual privacy of the farmer while providing a full range of data utility to the researcher. 

\vspace{1em}
\section{Acknowledgments}

This research was supported in part by the National Science Foundation (NSF) under awards OAC-2112606 and 2112533. Also, this research was partly supported by the United States Department of Agriculture (USDA) under grant number NR233A750004G019. 

\vspace{1em}
\vspace{1em}

%\printbibliography{}
%
\setcounter{secnumdepth}{0}  

%\section{Appendix}
% \subsection{Reproducibility Checklist:}
%\input{Sections/ReproducibilityChecklist}

\bibliographystyle{unsrt}  
\bibliography{references}  %%% Remove comment to use the external .bib file (using bibtex).

\end{document}